\documentclass[conference]{IEEEtran}
\IEEEoverridecommandlockouts
\usepackage{cite}
\usepackage{amsmath,amssymb,amsfonts}
\usepackage{algorithmic}
\usepackage{graphicx}
\usepackage{textcomp}
\usepackage[dvipsnames]{xcolor}
\usepackage{makecell}
\usepackage{multirow}
\usepackage{microtype}
\usepackage{array}
\usepackage{booktabs}
\usepackage{marvosym}

\def\BibTeX{{\rm B\kern-.05em{\sc i\kern-.025em b}\kern-.08em
    T\kern-.1667em\lower.7ex\hbox{E}\kern-.125emX}}
\begin{document}

\title{UFO: Unified Fact Obtaining for Commonsense Question Answering
\thanks{\textsuperscript{\Letter}Corresponding author.}
\thanks{The research is supported by the National Key R\&D Program of China (2020YFB1313601) and the National Science Foundation of China (No.62076174, No.61836007).}}

\author{
\IEEEauthorblockN{Zhifeng Li}
\IEEEauthorblockA{
\textit{School of Computer Science and Technology} \\
\textit{Soochow University}, Suzhou, China \\
li\_2hifeng@outlook.com}
\\
\IEEEauthorblockN{Yifan Fan}
\IEEEauthorblockA{
\textit{School of Computer Science and Technology} \\
\textit{Soochow University}, Suzhou, China \\
yifanfannlp@gmail.com}
\and 
\IEEEauthorblockN{Bowei Zou}
\IEEEauthorblockA{
\textit{Institute for Infocomm Research} \\
\textit{A*STAR}, Singapore \\
zou\_bowei@i2r.a-star.edu.sg}
\\
\IEEEauthorblockN{Yu Hong\textsuperscript{\Letter}}
\IEEEauthorblockA{
\textit{School of Computer Science and Technology} \\
\textit{Soochow University}, Suzhou, China \\
tianxianer@gmail.com
}}

\maketitle

\begin{abstract}
Leveraging external knowledge to enhance the reasoning ability is crucial for commonsense question answering. However, the existing knowledge bases heavily rely on manual annotation which unavoidably causes deficiency in coverage of world-wide commonsense knowledge. Accordingly, the knowledge bases fail to be flexible enough to support the reasoning over diverse questions. Recently, large-scale language models (LLMs) have dramatically improved the intelligence in capturing and leveraging knowledge, which opens up a new way to address the issue of eliciting knowledge from language models. We propose a Unified Facts Obtaining (UFO) approach. UFO turns LLMs into knowledge sources and produces relevant facts (knowledge statements) for the given question. We first develop a unified prompt consisting of demonstrations that cover different aspects of commonsense and different question styles. On this basis, we instruct the LLMs to generate question-related supporting facts for various commonsense questions via prompting. After facts generation, we apply a dense retrieval-based fact selection strategy to choose the best-matched fact. This kind of facts will be fed into the answer inference model along with the question. Notably, due to the design of unified prompts, UFO can support reasoning in various commonsense aspects (including general commonsense, scientific commonsense, and social commonsense). Extensive experiments on CommonsenseQA 2.0, OpenBookQA, QASC, and Social IQA benchmarks show that UFO significantly improves the performance of the inference model and outperforms manually constructed knowledge sources.


\end{abstract}


\section{Introduction}

As common sense presents the universal human consensus about the same thing and plays a critical role in daily life, there have been many efforts made on commonsense question answering (CQA) in the NLP community recently. A variety of CQA datasets have been published, which focus on different types of commonsense. For instance, CommonsenseQA 2.0~\cite{talmor2021commonsenseqa}  (CSQA2) possesses general commonsense, while Social IQA~\cite{sap-etal-2019-social} (SIQA) comprises social commonsense. Besides, both QASC~\cite{khot2020qasc} and OpenBookQA~\cite{mihaylov2018can} (OBQA) are primarily consisted of scientific commonsense.

The commonality of the aforementioned datasets is that a CQA model is required to understand and reason with the underlying commonsense knowledge before tackling any question in them. More seriously, the commonsense knowledge is neither explicitly presented in the question nor ever mentioned. Utilizing external knowledge sources to support models in commonsense reasoning is an effective way to address this issue. Previous approaches~\cite{li2021winnowing,feng2020scalable} applied multiple manually-constructed knowledge sources (e.g., ConceptNet~\cite{bosselut2019comet}, Wikipedia\footnote{https://wikipedia.org/}, Open Book~\cite{mihaylov2018can}, ATOMIC~\cite{hwang2021comet}, and Wiktionary\footnote{https://www.wiktionary.org/}) to the CQA task, which have successfully improved the performance of current CQA models to some extent.

\begin{table}[t]
\caption{A Commonsense Question and Related External Knowledge.}
\begin{center}
\begin{tabular}{|c|c|}
\hline
\textbf{Question} & \makecell[l]{
\specialrule{0em}{0.5pt}{0.5pt}
\textit{If none of the \textcolor{blue}{chickens} were \textcolor{orange}{males}, the chickens} \\
\textit{would still \textcolor{Green}{lay eggs}. Is it true?}\\
\specialrule{0em}{0.5pt}{0.5pt}
}\\
\hline
\makecell[c]{ATOMIC} & \makecell[l]{
\specialrule{0em}{0.5pt}{0.5pt}
\textit{PersonX fees the \textcolor{blue}{chickens}} -- \textit{Effects on PersonX} \\
-- \textit{As a result, PersonX wants harvest \textcolor{Green}{eggs}.} \\
\specialrule{0em}{0.5pt}{0.5pt}
} \\
\hline
\makecell[c]{Wikipedia } & \makecell[l]{
\specialrule{0em}{0.5pt}{0.5pt}
\textit{In the High Middle Ages, \textcolor{blue}{chickens} became less} \\ \textit{aggressive and began to \textcolor{Green}{lay eggs} earlier in the} \\ 
\textit{breeding season.} \\
\specialrule{0em}{0.5pt}{0.5pt}
} \\

\hline
\makecell[c]{ConceptNet} & \makecell[l]{
\specialrule{0em}{0.5pt}{0.5pt}
\textit{Hen} -- \textit{Is a type of} -- \textit{\textcolor{blue}{Chicken}} \\
\textit{Hen} -- \textit{Is a type of} -- \textit{\textcolor{orange}{female}} \\
\textit{Hen} -- \textit{Is capable of} -- \textit{\textcolor{Green}{Lay eggs}} \\
\specialrule{0em}{0.5pt}{0.5pt}
} \\
\hline
\makecell[c]{GPT3+UFO\\(ours)} & \makecell[l]{
\specialrule{0em}{0.5pt}{0.5pt}
\textit{Female \textcolor{blue}{chickens} are capable of \textcolor{Green}{laying eggs} with-}\\
\textit{out the presence of a \textcolor{orange}{male}. However, the eggs} \\ 
\textit{will not be fertilized and will not hatch.} \\ 
\specialrule{0em}{0.5pt}{0.5pt}
}\\
\hline

\multicolumn{2}{l}{
\makecell[l]{
\\
The question is selected from the CommonsenseQA 2.0 dataset,\\
and the knowledge is acquired from different sources. Where the\\
same color represent the same type of concept. ``-- --'' represents \\
edges in knowledge graph.}}
\end{tabular}
\label{tab1}
\end{center}
\end{table}

However, these knowledge sources suffer from two unavoidable weaknesses: limited domain and information sparsity. Let's discuss the weakness using the CQA example in Table~\ref{tab1}, where we randomly pick up a commonsense question and retrieve knowledge from three manually-constructed knowledge sources (ATOMIC, Wikipedia and ConceptNet). It can be found in this case that, firstly, most knowledge sources are constrained in independent domains, possessing less generalizability. Specifically, ATOMIC is concerned with the relationship between social events. It does not fit the general commonsense question. Secondly, the information sparsity problem occurs in both free-text form and graph-form knowledge sources as follows: 
\begin{itemize}
    \item On one hand, knowledge sources in the free-text form like Wikipedia fail to provide relationships among concepts. 
    \item On the other hand, knowledge graphs establish inter-concept connections with vertices and edges. However, the connections are signaled by only a few pre-defined types, where a large number of relation types are omitted. 
\end{itemize}

Due to the information sparsity mentioned above, although the knowledge retrieved from Wikipedia (in Table~\ref{tab1}) refers to ``\textit{chicken}'' and ``\textit{lay eggs}'', it still ignores the concept ``\textit{male}'' in question. Similarly, ConceptNet provides the information on gender, i.e., ``\textit{Hen is a type of female and chicken}'' (in Table~\ref{tab1}), it does not cover the crucial clue of ``{\em a hen can lay eggs without a rooster}'' that is required for reasoning in CQA.

The two weaknesses mentioned above lead to a gap between the retrieved knowledge and the question context -- neither irrelevant to the question nor sufficiently detailed to support the reasoning. To alleviate this issue, previous studies~\cite{hwang2021comet, wang2020connecting} applied language models to knowledge generation via fine-tuning models in knowledge graphs. Benefiting from information captured during pre-training, language models can produce some reasonable paths that do not exist in the knowledge graph. Besides, some prompt-based~\cite{liu2023pre} methods~\cite{liu2022generated, wei2022chain} leverage the original knowledge of language models. This enables the generation of knowledge statements or reasoning paths. However, these approaches are task-specific and thus deficient in generalizability to different CQA datasets.

Accordingly, we suggest that a general knowledge acquisition method is crucial for CQA tasks, and more importantly, it necessarily enables the retrieved knowledge information to adapt to various domains of commonsense knowledge. Therefore, we propose a Unified Fact Obtaining (UFO for short) approach\footnote{Access code and data via https://github.com/Zaaachary/CSQA-UFO} to enhance CQA models. UFO utilizes language models to generate facts that are not only related to the question but are domain-consistent with knowledge sources. On this basis, UFO leverages the generated facts to strengthen commonsense understanding during encoding so as to provide interpretable clues for commonsense reasoning.

Structurally, UFO is grounded on a pipeline framework consisting of unified facts generation, dense retrieval-based selection, fact integrated answer inference. For fact generation, we develop a unified few-shot~\cite{brown2020language} prompt containing various fact generation demonstrations that cover different aspects of commonsense and different types of question styles (assertion judgment, regular question, sentence completion, question with context). In terms of this prompt, UFO can produce question-related supporting facts for various CQA datasets instead of a single dataset. Considering that there are some noise items in the generated facts, we apply a dense retrieval technique~\cite{karpukhin2020dense} to select the best item for the encoding and reasoning. Finally, we apply the generated facts to multiple CQA benchmarks by fine-tuning the state-of-the-art model. The experimental results show that UFO yields 6.5\%, 7.0\%, 10.4\%, and 1.4\% improvements in CSQA2, OBQA, QASC, and SIQA, respectively. Our contributions are as follows:
\begin{itemize}
    \item We utilize the knowledge acquisition and reasoning capabilities of LLMs to produce relevant and adaptable supporting facts for questions, thereby addressing the limitations of manually constructed knowledge sources.
    \item We develop a unified few-shot prompt to instruct the generation of LLMs. This prompt covers a wide range of commonsense aspects and different types of questions, thus allowing for generalization across various CQA tasks.
    \item We introduce a dense retrieval-based relevance assessment to select the most relevant fact, which helps to reduce the noise issue of generated knowledge.
\end{itemize}

The rest of the paper is organized as follows. Section II overviews the related work. Section III presents the technical details of UFO. Section IV introduces the experimental settings and main test results, as well as a series of auxiliary experiments. Section V makes the conclusion.

\section{Related Works}

\subsection{Utilizing External Knowledge Sources}

Several previous works leverage the external commonsense knowledge sources to improve the performance of models on various CQA tasks~\cite{talmor-etal-2019-commonsenseqa, talmor2021commonsenseqa, khot2020qasc, mihaylov2018can}. The utilization of external knowledge can be divided into explicit input and implicit injection. The former first retrieves commonsense knowledge from knowledge sources based on the question and then enables models to infer the answer by incorporating the knowledge. It sometimes introduces specific network structures to enhance the model's inference capability. For example, Lin et al.~\cite{lin2019kagnet} introduce ConceptNet, Li et al.~\cite{li2021winnowing} combine Open Mind Common Sense, Chen et al.~\cite{chen2020improving} adapt Cambridge Dictionary, Xu et al.~\cite{xu2021human} utilize the knowledge from other question answering datasets. The latter leverages pre-training or domain adaptation to captures more commonsense knowledge, thus enhancing the commonsense reasoning capability of models. However, these manually constructed knowledge sources suffer from the problem of knowledge sparsity, thus unable to flexibly provide knowledge for various questions. To address this issue, we transform large-scale language models into knowledge sources, which brings the benefit of knowledge scalability.

\subsection{Eliciting Knowledge from Language Models}

Large-scale language models have learned amount of knowledge during the pre-training stage, and many works leverage these knowledge from the model via generative methods. For instance, fine-tuning generative models on knowledge graphs allows models to dynamically generate concept paths~\cite{wang2020connecting} or related events/entities~\cite{hwang2021comet} according to the input. Besides, some prompting-based method~\cite{paranjape-etal-2021-prompting,liu2022generated} guide models to produce explanation or related knowledge for each dataset. However, these approaches are task-specific and cannot be applied to different aspects of commonsense questions. Thus, from the outset of method development, we set generalization in CQA tasks as the primary goal, as commonsense questions in the real world have different styles of asking and covering various aspects of commonsense knowledge.

\section{APPROACH}
\begin{figure}[tp]
\centerline{\includegraphics{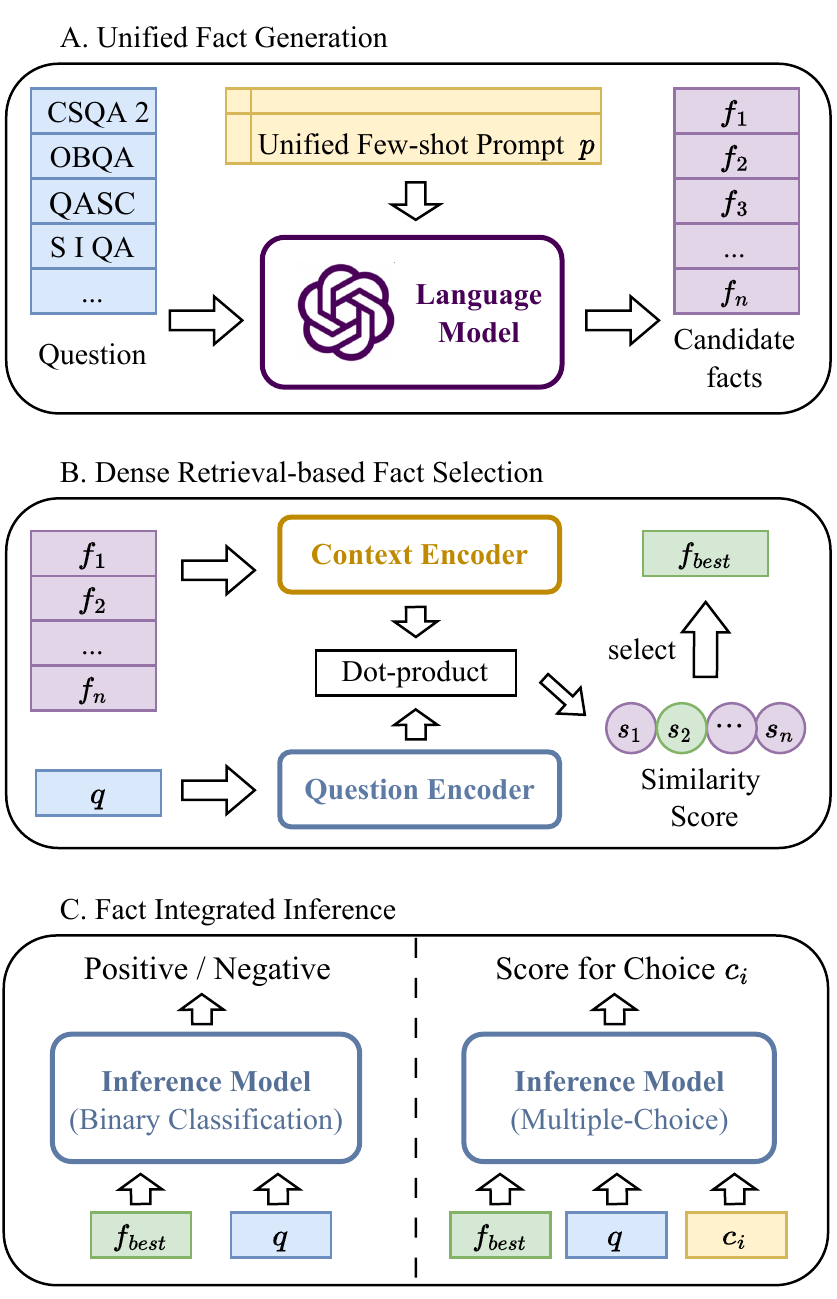}}
\caption{The overall architecture of Unified Fact Obtaining approach.}
\label{fig1}
\end{figure}

As illustrated in Fig.~\ref{fig1}, the proposed UFO approach consists of unified fact generation, dense retrieval-based selection, and fact integrated inference. Specifically, given a commonsense question $q$, we adapt it to the prompt and feed them into the fact generation model to sample $n$ candidate facts $F = \{f_1, f_2, ..., f_n\}$, where $f_i$ denotes the $i$-th fact item. Subsequently, we assess the similarity $S = \{s_1, s_2, ..., s_n\}$ between the question and each candidate fact via a dense retrieval technique and select the best fact $f_{best}$ with highest score. Finally, we conduct fact integrated inference on CQA tasks, including binary classification and multiple-choice.

\subsection{Unified Fact Generation}
We apply the few-shot prompting~\cite{pmlr-v139-zhao21c} method to leverage the commonsense knowledge reserve and language generation capabilities of large-scale language models, such as GPT-3~\cite{brown2020language}, LaMDA~\cite{thoppilan2022lamda}, and PaLM~\cite{chowdhery2022palm}. Unlike fine-tuning, few-shot prompting does not modify the model's parameters, thus preserving the pre-training gained knowledge. To facilitate the understanding of our approach, we first introduce the few-shot prompting method and then present the construction details of the unified fact generation prompt.

\begin{table*}[t]
\caption{Unified Prompt for Facts Generation.}
\begin{center}
\renewcommand{\arraystretch}{1}
\begin{tabular}{|c|c|}
\hline
\textbf{Part} & \textbf{Prompt Text}\\
\hline
\makecell[c]{\textbf{Head Instruction}} & \makecell[l]{
\specialrule{0em}{0.5pt}{0.5pt}
\textit{Relevant facts can be beneficial in question answering or assertion judgment, here are some examples:}\\
\specialrule{0em}{0.5pt}{0.5pt}
}\\
\hline
\makecell[c]{\textbf{CSQA2}\\(Positive)} & \makecell[l]{
\specialrule{0em}{0.5pt}{0.5pt}
Input: \textbf{Some Arizona cities have over a million people.} \\
Fact: \textit{Arizona is the 14th most populous state in the US, with over 7.3 million people. Of those, the cities}\\
\textit{of Phoenix, Tucson, Mesa, and Chandler each have populations of over a million people.}\\
\specialrule{0em}{0.5pt}{0.5pt}
}\\
\hline
\textbf{SIQA} & \makecell[l]{
\specialrule{0em}{0.5pt}{0.5pt}
Input: \textbf{Jordan was in charge of taking the food on the camping trip and left all the food at home.}\\ \textbf{How would Jordan feel afterwards?}\\
Fact: \textit{Forgetting to take the food for the camping trip is a failure of responsibility. A person may feel} \\
\textit{embarrassed and regretful after they have failed to fulfill their duty.}\\
\specialrule{0em}{0.5pt}{0.5pt}
}\\
\hline
\textbf{OBQA} & \makecell[l]{
\specialrule{0em}{0.5pt}{0.5pt}
Input: \textbf{Small reptiles in Texas can be brown or green on command, we call this what?}\\
Fact: \textit{Texas horned lizards are small reptiles native to Texas and other parts of the southern US. They}\\
\textit{have the ability to change their colour on command in order to camouflage with their environment.}\\
\specialrule{0em}{0.5pt}{0.5pt}
}\\
\hline
\makecell[c]{\textbf{CSQA2}\\(Negative)} & \makecell[l]{
\specialrule{0em}{0.5pt}{0.5pt}
Input: \textbf{There is at least one example of human blood type in the following list: AC, AH, C, OH, BB.}\\
Fact: \textit{There are 4 different blood types –A, B, AB and O. These names indicate whether the blood’s red}\\
\textit{cells carry the A antigen, the B antigen, both A and B antigens, or neither antigen.}\\
\specialrule{0em}{0.5pt}{0.5pt}
}\\
\hline
\textbf{QASC} & \makecell[l]{
\specialrule{0em}{0.5pt}{0.5pt}
Input: \textbf{The interior chambers have tiny what that trap the particles.}\\
Fact: \textit{In air purifiers, the interior chambers contain filters made of fibres, such as activated carbon, that}\\
\textit{are designed to trap particles such as dust, pollen, and other allergens.}\\
\specialrule{0em}{0.5pt}{0.5pt}
}\\
\hline
\makecell[c]{\textbf{Tail Instruction}} & \makecell[l]{
\specialrule{0em}{0.5pt}{0.5pt}
\textit{According to above examples, fact generation requires a descriptive knowledge statement related to the}\\
\textit{given input rather than a direct answer or judgment. Now, generate relevant facts (30 words or less)}\\
\textit{for new input.}\\
\specialrule{0em}{0.5pt}{0.5pt}
}\\
\hline
\makecell[c]{\textbf{Placeholder}} & \makecell[l]{
\specialrule{0em}{0.5pt}{0.5pt}
Input: \{question\} \\
Fact:\\
\specialrule{0em}{0.5pt}{0.5pt}
}\\
\hline

\multicolumn{2}{l}{
\makecell[l]{
\\
To facilitate the presentation, we split the prompt into several part, including head, demonstrations from different datasets and\\tail. There are two line breaks between each part to indicate the segment.}}
\end{tabular}
\label{tab2}
\end{center}
\end{table*}

\subsubsection{Few-shot Prompting}
LLMs have demonstrated an impressive capacity for reasoning and pattern recognition. Based on these powerful abilities, few-shot prompting, also known as in-context learning, makes pre-trained LLMs adaptable to various types of tasks without extra training. Given a prompt consisting of a few task demonstrations (input-output pairs) and an input to be predicted, models can generate the corresponding output based on the patterns recognized from previous samples. For instance, when presented with a prompt of English-French translation pairs such as ``$En_1$-\textgreater $Fr_1$, $En_2$-\textgreater $Fr_2$, ..., $En_n$-\textgreater'', where ``$En_i$-\textgreater$Fr_i$'' denotes the $i$-th English-French pair, the model can infer that it should generate the French translation of English sentence ``$En_n$'' next.

For fact generation, we apply few-shot prompting to instruct the model producing question-related fact (knowledge statement). Formally, given a commonsense question $a$, we plug it into the placeholder of prompt and then obtain the input sequence $pq$. Next, we feed $pq$ into a LLM and sample $n$ different fact outputs $F = \{f_1, f_2, ..., f_n\}$.

\subsubsection{Question Selection}
We develop a unified few-shot prompt that guides the LLMs to generate relevant facts for various commonsense questions. As shown in Table~\ref{tab2}, our unified prompt consists of a head instruction, several demonstrations of question-fact pairs, a tail instruction, and a placeholder for the new question. The most important part of the prompt is the demonstration of question-fact pairs. To diversify the demonstrations, we collected questions from four different CQA datasets (CSQA2, SIQA, OBQA, and QASC), which cover different knowledge aspects such as general, scientific, and social commonsense. These questions also encompass the major styles of asking questions, including regular questions (OBQA), assertion judgments (CSQA2), sentence completions (QASC), and questions with context (SIQA). Notably, the prompt contains one question for each type of style, except for assertion judgments, which take two places (positive and negative). This is because assertions can be either positive (correct) or negative (incorrect), if only the correct case is included, models will be mislead to support the assertion-form input regardless of its correctness. 

\subsubsection{Fact Annotation}
After collecting questions, we annotate the demonstration facts for each question, ensuring that the facts adhered to three principles: (1) facts have to be objective and correct. (2) facts should directly or potentially support the reasoning of the question. (3) facts have to consist of two parts, one-step and two-step extension of the concepts mentioned in the question, thus simulating a multi-hop reasoning process. For example, we take the positive sample of CSQA2 ``\textit{Some Arizona cities have over a million people.}'' The annotated facts start from Arizona's population, which connects the concepts ``\textit{Arizona}'' and ``\textit{people}'' in question. Then we establish the connection between ``\textit{Arizona cities}'' and ``\textit{a million people}'' by listing cities that satisfied the condition.

\subsubsection{Head and Tail Instructions}
Building on prior researches~\cite{liu2022generated,brown2020language}, we included an instruction at the start of the prompt to define the relationship between the input (question) and output (fact). However, we find that models sometimes provide direct answer to the given question rather than generating supporting facts. To address this issue, we added an additional tail instruction between the demonstrations and the question placeholder. The tail instruction summarizes the characteristics of the fact, limits the word count, and asks models not to generate direct answers.

\subsection{Dense Retrieval-based Fact Selection}
Despite the strength of the LLMs, the presence of incorrect or irrelevant noisy items in generated facts is unavoidable, particularly when posing challenging questions or employing models with fewer parameters. To address this noise problem, we introduce Dense Passage Retrieval~\cite{karpukhin2020dense} (DPR), an off-the-shelf method for retrieving relevant information from large amounts of text. DPR, based on a dual-encoder architecture, consists of a question encoder and a context encoder. Following the primary usage\footnote{{https://huggingface.co/docs/transformers/model\_doc/dpr}}, we encode the commonsense question $q$ and each candidate fact $f_i$ into dense vector representations. Then, we compute the dot-product similarity $S$ between the question and each fact. Finally, the best fact item with the highest score is adopted for the inference model.

%


\subsection{Facts Integrated Inference}
We select DeBERTa-v3~\cite{he2021debertav3} as the inference model, and fine-tuned it on CQA tasks by minimizing the cross-entropy loss. This enabled the model to predict the answer through reasoning over the question and related facts. The inference model first encodes the given input (the best fact and QA sample) and then outputs the representation of the ``[CLS]'' token $h$ (input representation). The task form of CQA can be divided into two categories, namely multiple-choice and binary classification, which necessitate distinct inputs and different usages of the input representation $h$.

For the binary classification task, like CSQA2, we concatenate the best fact $f_{best}$ and the question $q$ in the form of ``[CLS] $f_{best}$ [SEP] $q$ [SEP]''. We then feed this sequence into the inference model to obtain the representation $h$, which is further reduced to a 2-dimensional vector $l$ via a feed-forward network (FFN). Finally, we normalize $l$ with a softmax function to obtain the probabilities of positive and negative categories.

For multiple-choice tasks such as OBQA, each choice is encoded separately. For the $i$-th choice $c_i$, it is combined with $q$ and $f_{best}$ in the format of ``[CLS] $f_{best}$ [SEP] $q$ [SEP] $c_i$ [SEP]''. Then, the inference model encodes the input sequence to obtain $h_i$. Unlike binary classification tasks, the FFN projects $h_i$ into a 1-dimensional scalar $p_i$, which represents the plausibility of choice $c_i$. The plausibility of all $k$ choices $P = \{p_1, p_2, ..., p_k\}$ is then computed and the final probability of all choices is obtained via softmax. Finally, the candidate answer with the highest probability is considered the predicted result.
\section{Experiment}

\subsection{Datasets and Compared Methods}
To assess the applicability of our unified facts obtaining approach, we present four commonsense question answering datasets that focus on distinct commonsense aspects. All of these datasets use Accuracy as the evaluation metric, which is calculated as the ratio of correctly answered samples to the total number of samples. A brief introduction of each dataset and corresponding compared methods are listed below.

\textbf{CSQA2}~\cite{talmor2021commonsenseqa} focuses on the general commonsense, including assertion judgment and yes/no question answering sample, which can be considered a positive/negative binary classification dataset. To compare our method with large-scale pre-trained language models, we select three models, namely T5-11B~\cite{raffel2020exploring}, UNICORN~\cite{lourie2021unicorn}, and UL-20B~\cite{tay2022unifying}. Where UNICORN is a T5-11B model that has been fine-tuned on the RAINBOW commonsense corpus, thus providing an advantage when applied to commonsense reasoning tasks.

\textbf{OBQA}~\cite{mihaylov2018can} models the scenario of open book examination for assessing the understanding of a subject and involves elementary-level science questions. Each question in OBQA has four candidate answers. OBQA provides an open-book corpus, which shares the same domain and has a large overlap with OBQA. We introduce T5-11B + KB and Unified QA (T5-11B) for comparison since they utilize the in-domain knowledge from open book corpus. Besides, we also compare with AristoRoBERTa + GSC~\cite{wang2021gnn}, since it leverages the concept relation information from the knowledge graph and applies a graph neural module to process the graph information.

\textbf{QASC}~\cite{khot2020qasc} is an 8-way multiple-choice dataset focusing on sentence composition of grade school science. In addition to the question and candidate answers, QASC also provides a corpus of 17M sentences (in-domain data). We select RoBERTa + KF + SIR v2~\cite{banerjee2020knowledge} and RoBERTa + AIR~\cite{yadav-etal-2020-unsupervised} for comparison, both of which incorporate the in-domain knowledge. The former consists of a knowledge re-ranking model and knowledge fusion inference model, while the latter applies an iterative multi-hop knowledge retrieval method.

\textbf{SIQA}~\cite{sap-etal-2019-social} is a benchmark for testing social commonsense intelligence, focuses on reasoning about the actions and social implications of people. Each sample in SIQA has 3 candidate answers. Beside unified QA (T5-11B), we also add RoBERTa + ATOMIC for comparison. This method introduces the ATOMIC, which is a commonsense knowledge graph containing physical entities, social events, and interactions. Notably, ATOMIC is used to construct SIQA, so it is an in-domain knowledge source.

\subsection{Implementation Details}

For unified fact generation, we chose GPT-3 Davinci (175B) as the mainstay of our experiment, as it is one of the most powerful models that can be used in few-shot settings. And we also considered GPT-neo~\cite{gpt-neo} (2.7B) and GPT-3 Curie (6.7B) as the facts generation models. For GPT-3 Davinci, we generate $n=3$ facts for each question, while for the GPT3 Curie and GPT-neo, we set $n=5$. The generation is based on nucleus sampling with top-$p=0.5$ and $temperature=0.7$.

For fact integrated inference, we chose DeBERTa-v3~\cite{he2021debertav3} Large (418M) as our backbone model. We fine-tuned the model using the AdamW optimizer~\cite{loshchilov2017decoupled}. We conduct hyperparameter grid-search for each dataset, and the best performing models are selected for predictions on the test set. We choose the weight decay in \{0, 0.01\}. The batch size is chosen from the set \{8, 16, 32\} and the learning rate is chosen from the range \{3e-6, 5e-6, 7e-9, 9e-6, 1e-5\}.

\subsection{Experiment Results and Analysis}

\begin{table}[tb]
\caption{Performance Comparison on Four CQA Benchmarks.}
\begin{center}
\renewcommand{\arraystretch}{1.1}
\begin{tabular}{|c|l|c|c|}
\hline
\textbf{Dataset} & \textbf{Models} & \textbf{\makecell[c]{Acc\\(dev)}} & \textbf{\makecell[c]{Acc\\(test)}}\\
\hline
\multirow{5}*{CSQA2} & DeBERTa & 69.5 & 64.3 \\
~ & T5-11B & 68.5 & 67.4 \\
~ & UNICORN (T5-11B) & 69.9 & 69.6 \\
~ & UL-20B & / & 70.1 \\
~ & \textbf{DeBERTa + UFO (ours)} & \textbf{73.9} & \textbf{70.8} \\
\hline
\multirow{5}*{OBQA} & DeBERTa & 82.2 & 82.6 \\
~ & T5-11B + KB * & / & 85.4 \\
~ & Unified QA (T5-11B) * & / & 87.2 \\
~ & AristoRoBERTa + GSC & / & 87.4\\
~ & \textbf{DeBERTa + UFO (ours)} & \textbf{89.8} & \textbf{89.6} \\
\hline
\multirow{4}*{QASC} & DeBERTa & 75.4 & 72.3 \\
~ & RoBERTa + KF + SIR v2 * & \textbf{85.2} & 80.4 \\
~ & RoBERTa + AIR * & / & 81.4 \\
~ & \textbf{DeBERTa + UFO (ours)} & 84.5 & \textbf{82.7} \\
\hline
\multirow{4}*{SIQA} & DeBERTa & 81.9 & 80.6 \\
~ & RoBERTa + ATOMIC * & / & 81.0 \\
~ & Unified QA (T5-11B) & / & 81.5 \\
~ & \textbf{DeBERTa + UFO (ours)} & 83.1 & \textbf{82.0} \\
\hline
\multicolumn{4}{l}{
\makecell[l]{
\\
The asterisk ``*'' denotes models that have been integrated\\
with in-domain knowledge sources. The test scores come\\
from the corresponding leaderboard.\\
}}
\end{tabular}
\label{tab3}
\end{center}
\end{table}

\subsubsection{Main Experiment}
Table~\ref{tab3} illustrates the performance of our UFO approach and other compared methods on four CQA benchmarks. From our observations, we conclude:
The UFO method has been demonstrated to be superior to its respective DeBERTa baselines on four datasets, with improvements of 6.5\%, 7.0\%, 10.4\%, and 1.4\%, respectively. This indicates that UFO can effectively generate commonsense knowledge on different aspects and can be applied to various CQA datasets. On CSQA2, QASC and SIQA datasets, the DeBERTa baseline model encountered severe overfitting problems, with an absolute dev-test gap of 5.2\%, 3.1\% and 1.3\%, respectively. UFO not only improved the performance but also narrowed this dev-test gap to 3.1\%, 1.8\% and 1.0\%, respectively, indicating that UFO brings better generalization capabilities.

The OBQA and QASC datasets provide in-domain scientific corpus, which have a significant overlap with the question samples. We annotate methods imported in-domain corpus with an asterisk (*). Compared with these method, our approach achieves higher accuracy, which indicates that UFO is capable of flexibly adapting to multiple knowledge datasets and generating high-quality knowledge. 

Our approach yields improvements on the SIQA dataset, although the degree of improvement was not as significant as with other datasets. This is because the SIQA dataset is designed to evaluate the ability to infer human behavior in social situations, while our approach provides explanations or descriptions of the given scenario, mainly providing potential assistance for reasoning.

On CSQA2, the UL model introduces 20B parameters, which capture more commonsense knowledge in the pre-training stage. While the UNICRON approach implicitly injects external knowledge to T5 through domain adaption. In contrast, our proposed UFO approach achieves a higher accuracy (70.8\%) through direct knowledge input, indicating that explicit knowledge input is more effective than implicit knowledge injection in terms of knowledge utilization.

\subsubsection{Ablation Study}
We conducted ablation experiments on the OBQA dataset using three fact generation models of different scales, namely GPT-Neo (2.7B), GPT-3 Curie (6.7B) and GPT-3 Davinci (175B). The results of the experiments are presented in Table~\ref{tab4}. We can observe that the combination of largest generation model and dense retrieval-based fact selection yields the highest improvement on the OBQA dataset. And the results of the two groups of experiments suggest that fact selection is effective when using models of various sizes. Furthermore, comparison on fact-generation models in different sizes reveals that the accuracy of the inference model increases proportionally with the increase in parameters of the fact generation model. This indicates that the useful information covered by the facts is positively correlated with the scale of the generation model. Notably, even when relatively small models are adopted for generation, the produced facts still have a positive effect on question answering.

\begin{table}[t]
\caption{Ablation Study on OpenBookQA Benchmark.}
\begin{center}
\renewcommand{\arraystretch}{1.1}
\begin{tabular}{|c|l|c|}
\hline
\multicolumn{2}{|l|}{\textbf{Models}} & \textbf{Acc (dev)}\\
\hline
\multicolumn{2}{|l|}{DeBERTa Baseline} & 82.2\\
\hline
 \multirow{3}*{w/o Selection} & + Fact (2.7B, GPT-Neo) & 83.0\\
 ~ & + Fact (6.7B, GPT-3 Curie) & 83.1\\
 ~ & + Fact (175B, GPT-3 Davinci) & 89.2\\
 \hline
\multirow{3}*{w/ Selection} & + Fact (2.7B, GPT-Neo) & 83.8\\
 ~ & + Fact (6.7B, GPT-3 Curie) & 84.5\\
 ~ & + Fact (175B, GPT-3 Davinci) & \textbf{89.8}\\
\hline
\end{tabular}
\label{tab4}
\end{center}
\end{table}

\subsubsection{Comparison of Reasoning Ability}
Table~\ref{tab4.1} presents the performance of backbone models involved in the UFO approach (DeBERTa and GPT-3), and the advanced large model GPT-3.5-turbo, when answering questions on the QASC dataset. The prompt for zero-shot question answering are list in Appendix A. Comparing the results of DeBERTa and GPT-3, it can be observed that due to its huge parameter size, GPT-3 captures rich commonsense information and achieves higher accuracy. On the other hand, GPT-3.5-turbo, also known as ChatGPT\footnote{https://openai.com/blog/chatgpt}, is fine-tuned with instructions and reinforcement learning~\cite{ouyang2022training} on the basis of GPT-3, thus possessing stronger commonsense reasoning ability and task adaptation ability. The UFO approach generates facts based on GPT-3 and utilizes DeBERTa with relatively smaller parameter size for reasoning, achieving better performance than both GPT-3 and GPT-3.5-turbo. This result demonstrates the efficiency of the pipeline architecture, where knowledge is generated first, followed by the answer inference.

\begin{table}[t]
\caption{Reasoning Ability Comparison on QASC Benchmark.}
\begin{center}
\renewcommand{\arraystretch}{1.1}
\begin{tabular}{|l|c|c|}
\hline
\textbf{Models} & \textbf{Model Scale} & \textbf{Acc (dev)}\\
\hline
DeBERTa & 418M & 75.4\\
GPT-3 Davinci (Zero-shot) & 175B & 77.5 \\
GPT-3.5-turbo (Zero-shot) & 175B & 80.0 \\
\hline
\textbf{DeBERTa + UFO (ours)} & 175B + 418M & \textbf{83.1} \\
\hline
\end{tabular}
\label{tab4.1}
\end{center}
\end{table}

\subsection{Qualitative Analysis} 

In order to evaluate the quality of knowledge produced by the model, we conducted a manual evaluation of the model-generated results. First, we sampled 25 samples from each of the four datasets, thus constructing an evaluation set of 100 samples. Then we used the UFO method to generate relevant facts for these samples. Finally, we manually evaluated these facts, categorizing them into directly helpful facts ``DH'', potentially helpful facts ``PH'', and unhelpful(irrelevant or erroneous) facts ``UH''. The statistics are presented in Table~\ref{tab5}. Overall, 64\% of the generated facts are directly helpful, and 24\% of them are potentially helpful, which indicates that the model-generated facts generally conform to human common sense. Furthermore, we discover that the highest proportions (96\%) of helpful facts (DH \& PH) are present on QASC and OBQA. In contrast, the proportion was slightly lower on CSQA2 and SIQA, and further demonstrates the difficulty of the individual datasets.

\begin{table}[tb]
\caption{Statistics on the Quality of Generated Facts.}
\begin{center}
\begin{tabular}{|c|ccccc|}
\hline
\textbf{Cat.} & \textbf{CSQA2} & \textbf{OBQA} & \textbf{QASC} & \textbf{SIQA} & \textbf{Overall}\\
\hline
DH & 15 (60\%) & 19 (76\%) & 17 (68\%) & 13 (52\%) & 64\%\\
PH & 4 (16\%) & 5 (20\%) & 7 (28\%) & 7 (28\%) & 23\%\\
UH & 6 (24\%) & 1 (4\%) & 1 (4\%) & 5 (12\%) & 13\%\\
\hline
\multicolumn{6}{l}{
\makecell[l]{
\\
The percentages in parentheses refer to each dataset.
}}
\end{tabular}
\label{tab5}
\end{center}
\end{table}

\begin{table}[t]
\caption{Cases from QASC and OpenBookQA.}
\begin{center}
\renewcommand{\arraystretch}{1}
\begin{tabular}{|l|}
\multicolumn{1}{l}{\makecell[l]{\textbf{Case \#1} QASC\\}}\\
\hline
Question: \textit{Where is the device to measure wind placed?}\\
Answer: A. \textit{At the top of a station}\\
\hline
\makecell[l]{
\textbf{Fact 1}:\\
\textit{The wind is measured by the speed of the wind.}\\
}\\
\hline
\makecell[l]{
\textbf{Fact 2} (selected by DPR):\\
\textit{The wind is measured by anemometer, which is placed on the roof}\\
\textit{of the building.}\\
}\\
\hline
\makecell[l]{
\textbf{Fact 3}:\\
\textit{The wind speed is measured by a device called an anemometer.}\\
}\\
\hline
\multicolumn{1}{l}{\makecell[l]{\\
\textbf{Case \#2} OBQA\\
}}\\
\hline
Question: \textit{\textcolor{blue}{Plant} requires what for \textcolor{blue}{photosynthesis}?}\\
Answer: B. \textit{\textcolor{orange}{light} from our closest star}\\
\hline
\makecell[l]{
\textbf{GPT-3 Davinci}:\\
\textit{\textcolor{blue}{Photosynthesis} is the process by which \textcolor{blue}{plants} use \textcolor{orange}{light} energy from}\\
\textit{the \textcolor{orange}{sun}, carbon dioxide from the air, and water to produce food.}\\
}\\
\hline
\makecell[l]{
\textbf{GPT-3 Curie}:\\
\textit{\textcolor{blue}{Plants} require \textcolor{orange}{light} to produce energy in the form of glucose from}\\
\textit{carbon dioxide and water.}\\
}\\
\hline
\makecell[l]{
\textbf{GPT-Neo}:\\
\textit{\textcolor{blue}{Plants} require \textcolor{orange}{sunlight} to produce carbohydrates, which they use to}\\
\textit{produce sugars.}\\
}\\
\hline

\multicolumn{1}{l}{
\makecell[l]{
\\
Facts in Case \#1 are sampled from GPT-Neo. And \textcolor{blue}{blue} denotes\\
entities from the question while \textcolor{orange}{orange} denotes the entities from\\
the correct answer.\\
}
}
\end{tabular}
\label{tab6}
\end{center}
\end{table}

\subsection{Case Study}

As shown in Table~\ref{tab6}, we select two cases to reveal the effect of the UFO approach. Case \#1 displays a question about the location of the wind measurement device from the QASC dataset. We utilize our unified prompt and sample three facts from GPT-Neo(2.7B). As we can see, Fact 1 is a wrong statement. Fact 3 refers to the ``anemometer'', which may potentially help judge, but not directly, and only Fact 2 mentions the location of the wind measurement device as ``roof of the building'', which increases the probability of the correct choice. This sample demonstrates the noise problem we encounter in fact generation. It is likely to pass on irrelevant or even wrong information to the inference model if not adequately handled. The introduction of the DPR-based selection alleviates this problem, like Case \#1, the most related fact is selected.

In Case \#2, we demonstrate a question from OBQA and the best facts produced by the UFO approach based on different fact generation models. In this question, it is essential  to understand the meaning of ``photosynthesis''. The best facts from the three models correctly explain photosynthesis and establish a connection with the ``light'' in the answer. Furthermore, we also observed that, as the number of parameters increases, the model's output is further enriched with more information.

\section{Conclusion}

In this paper, we proposed a Unified Facts Obtaining (UFO) approach for CQA task. UFO is capable of guiding large-scale language models to generate question-related supportive facts for various domains of questions, where a unified few-shot prompt is used. On the basis, we apply a dense retrieval-based fact selection strategy to choose the best fact from candidates, which relieves the interference of noise fact items. Extensive experiments on four benchmark CQA datasets demonstrate that UFO yields substantial improvements, and it outperforms the strong CQA models that were trained over the manually-constructed knowledge sources. More importantly, UFO can be used as a new baseline for further exploration of leveraging pre-trained language models, in the scenario of eliciting commonsense knowledge.

In the future, we will investigate the rewriting of questions using UFO (i.e., knowledge enhanced commonsense question rewriting). The rewriting model can be applied to augment the existing training data of CQA with the generalized languages. In addition, it increases the possibility that a sophisticated question generator cooperates with a simple reasoner and thus improves the efficiency of CQA models.  



\bibliographystyle{IEEEtran}
\bibliography{cite}

\newpage 
\appendix

\subsection{Prompts for Zero-shot Question Answering}

Table~\ref{tab8} demonstrates the prompt for GPT-3 and ChatGPT in Zero-shot Question Answering. Where GPT-3 has only one text domain, while, ChatGPT has three distinct domains: System, User, and Assistant. Where System domain pertains to the responses generated by the model for system-initiated prompts, such as those that provide information or instructions to the user. The user domain, meanwhile, deals with the model's responses to user-initiated prompts, such as questions or requests for assistance. Lastly, the assistant domain is responsible for generating responses that bridge the gap between the system and user domains, providing contextual information and guidance as needed. 

\begin{table}[h]
\caption{Prompt for Zero-shot Question Answering.}
\begin{center}
\renewcommand{\arraystretch}{1}
\begin{tabular}{|rl|}
\multicolumn{2}{l}{\makecell[l]{\textbf{GPT-3 Davinci}}}\\
\hline
\multicolumn{2}{|l|}{Select the best choice for the given question.}\\
\multicolumn{2}{|l|}{Question: \{question\}}\\
\multicolumn{2}{|l|}{Choices: A. choice; B. choice; C. choice ...}\\
\multicolumn{2}{|l|}{Answer:}\\
\hline
\multicolumn{2}{l}{\makecell[l]{\\
\textbf{ChatGPT (GPT-3.5-turbo)}\\
}}\\
\hline
Role & Content \\
\hline
System & Select the best choice for the given question.\\
User & \makecell[l]{Question: \{question\}\\
Choices: A. choice; B. choice; C. choice ...}\\
Assistant & Answer:\\
\hline
\end{tabular}
\label{tab8}
\end{center}
\end{table}

\end{document}